\documentclass[11pt]{article}
\usepackage{coling2018}
\usepackage{times}
\usepackage{url}
\usepackage{latexsym}

\usepackage{bm}
\usepackage{amsmath}
\usepackage{graphicx}
\usepackage{booktabs}

\makeatletter
\newcommand\figcaption{\def\@captype{figure}\caption}
\newcommand\tabcaption{\def\@captype{table}\caption}
\makeatother

\title{A Full End-to-End Semantic Role Labeler, \\Syntax-agnostic Over Syntax-aware?}

\author{Jiaxun Cai$^{1,2}$, Shexia He$^{1,2}$, Zuchao Li$^{1,2}$, Hai Zhao$^{1,2}$\thanks{ $\ \ $Corresponding author. This paper was partially supported by National Key Research and Development Program of China (No. 2017YFB0304100), National Natural Science Foundation of China (No. 61672343 and No. 61733011), Key Project of National Society Science Foundation of China (No. 15-ZDA041), The Art and Science Interdisciplinary Funds of Shanghai Jiao Tong University (No. 14JCRZ04).}\\
 $^{1}$Department of Computer Science and Engineering, Shanghai Jiao Tong University \\
 $^{2}$Key Laboratory of Shanghai Education Commission for Intelligent Interaction \\ and Cognitive Engineering, Shanghai Jiao Tong University, Shanghai, China\\
 {\tt \{caijiaxun, heshexia, charlee\}@sjtu.edu.cn, }
 \\ {\tt zhaohai@cs.sjtu.edu.cn }
}

\date{}

\begin{document}
\maketitle
\begin{abstract}
Semantic role labeling (SRL) is to recognize the predicate-argument structure of a sentence, including subtasks of predicate disambiguation and argument labeling. Previous studies usually formulate the entire SRL problem into two or more subtasks. For the first time, this paper introduces an end-to-end neural model which unifiedly tackles the predicate disambiguation and the argument labeling in one shot. Using a biaffine scorer, our model directly predicts all semantic role labels for all given word pairs in the sentence without relying on any syntactic parse information. Specifically, we augment the BiLSTM encoder with a non-linear transformation to further distinguish the predicate and the argument in a given sentence, and model the semantic role labeling process as a word pair classification task by employing the biaffine attentional mechanism. Though the proposed model is syntax-agnostic with local decoder, it outperforms the state-of-the-art syntax-aware SRL systems on the CoNLL-2008, 2009 benchmarks for both English and Chinese. To our best knowledge, we report the first syntax-agnostic SRL model that surpasses all known syntax-aware models.

\end{abstract}

\section{Introduction}
\blfootnote{
	This work is licensed under a Creative Commons 
	Attribution 4.0 International License.
	License details:
	\url{http://creativecommons.org/licenses/by/4.0/}
}
Semantic role labeling (SRL) is a shallow semantic parsing, which is dedicated to identifying the semantic arguments of a predicate and labeling them with their semantic roles. SRL is considered as one of the core tasks in the natural language processing (NLP), which has been successfully applied to various downstream tasks, such as information extraction \cite{Bastianelli2013}, question answering \cite{QA2007,berant2013}, machine translation \cite{MT2012,shi-EtAl:2016}. 

Typically, SRL task can be put into two categories: constituent-based (i.e., phrase or span) SRL and dependency-based SRL. This paper will focus on the latter one popularized by CoNLL-2008 and 2009 shared tasks \cite{surdeanu-EtAl2008,hajivc-EtAl2009}. Most conventional SRL systems relied on sophisticated handcraft features or some declarative constraints \cite{pradhan2005,Zhao2009Conll}, which suffers from poor efficiency and generalization ability. A recently tendency for SRL is adopting neural networks methods attributed to their significant success in a wide range of applications \cite{Bai2018deep,zhang2018OneShot}. However, most of those works still heavily resort to syntactic features. Since the syntactic parsing task is equally hard as SRL and comes with its own errors, it is better to get rid of such prerequisite as in other NLP tasks. Accordingly, \newcite{marcheggiani2017} presented a neural model putting syntax aside for dependency-based SRL and obtain favorable results, which overturns the inherent belief that syntax is indispensable in SRL task \cite{punyakanok-2008}. 

Besides, SRL task is generally formulated as multi-step classification subtasks in pipeline systems, consisting of predicate identification, predicate disambiguation, argument identification and argument classification. Most previous SRL approaches adopt a pipeline framework to handle these subtasks one after another. Until recently, some works \cite{zhou-xu2015,he-acl2017} introduce end-to-end models for span-based SRL, which motivates us to explore integrative model for dependency SRL.

In this work, we propose a syntactic-agnostic end-to-end system, dealing with predicate disambiguation and argument labeling in one model, unlike previous systems that treat the predicate disambiguation as a subtask and handle it separately. In detail, our model contains (1) a deep BiLSTM encoder, which is able to distinguish the predicates and arguments by mapping them into two different vector spaces, and (2) a biaffine attentional \cite{dozat2017deep} scorer, which unifiedly predicts the semantic role for argument and the sense for predicate. 

We experimentally show that though our biaffine attentional model remains simple and does not rely on any syntactic feature, it achieves the best result on the benchmark for both Chinese and English even compared to syntax-aware systems. In summary, our major contributions are shown as follows:
\begin{itemize}
\item We propose an accurate syntax-agnostic model for neural SRL, which outperforms the best reported syntax-aware model, breaking the long-held belief that syntax is a prerequisite for SRL.
\item Our model gives state-of-the-art results on the CoNLL-2008, CoNLL-2009 English and Chinese benchmarks, scoring 85.0\% F$_1$, 89.6\% F$_1$ and 84.4\% F$_1$, respectively.
\item Our work is the first attempt to apply end-to-end model for dependency-based SRL, which tackles the predicate disambiguation and the argument labeling subtasks in one shot.
\end{itemize}

\section{Semantic Structure Decomposition}
SRL includes two subtasks: predicate identification/disambiguation and argument identification/labeling. Since the CoNLL-2009 dataset provides the gold predicates, most previous neural SRL systems use a default model to perform predicate disambiguation and focus on argument identification/labeling. Despite nearly all SRL work adopted the pipeline model with two or more components, \newcite{Zhao2008Parsing} and \newcite{zhao-jair-2013} presented an end-to-end solution for the entire SRL task with a word pair classifier. Following the same formulization, we propose the first neural SRL system that uniformly handles the tasks of predicate disambiguation and argument identification/labeling.

\begin{figure}
	\centering
	\includegraphics[scale=0.85]{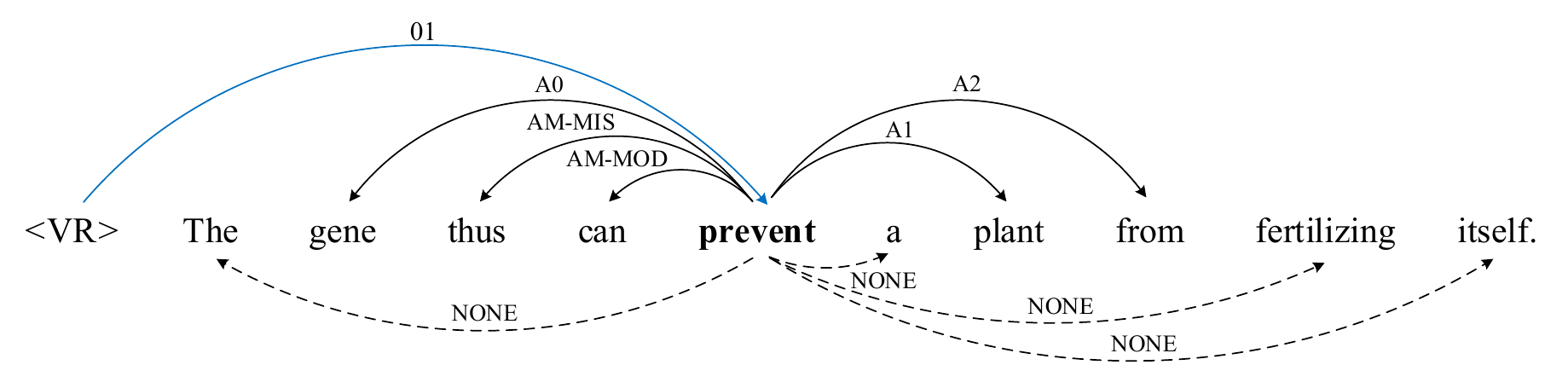}
	\caption{The dependency graph with a virtual node $<$VR$>$ and virtual arcs with \emph{None} label.} \label{dummy}
\end{figure}

In semantic dependency parsing, we can always identify two types of words,
semantic head (predicate) and semantic dependent (argument). To build the needed predicate-argument structure, the model only needs to predict the role of any word pair from the given sentence. For the purpose, an additional role label \emph{None} and virtual root node $<$\emph{VR}$>$ are introduced. The \emph{None} label indicates that there is no semantic role relationship inside the word pair. We insert a virtual root $<$\emph{VR}$>$ in the head of the sentence, and set it as the semantic head of all the predicates. By introducing the \emph{None} label and the $<$\emph{VR}$>$ node, we construct a semantic tree rooted at the $<$\emph{VR}$>$ node with several virtual arcs labeled with \emph{None}. Thus, the predicate disambiguation and argument identification/labeling tasks can be naturally regarded as the labeling process over all the word pairs. Figure \ref{dummy} shows an example of the semantic graph augmented with a virtual root and virtual arc, and Table 1 lists all the corresponding word pair examples, in which two types of word pairs are included, $<$\emph{VR}$>$ followed by predicate candidates\footnote{Note that there is a key difference between CoNLL 2008 and 2009 shared task for English, the latter has specified the predicate in the data so that here we have only one sample for $<$\emph{VR}$>$-predicate pair to make predicate disambiguation. For the former, predicate candidates should be every words in the given sentence. More details are seen in Section \ref{ablsec_conll08}.} and a known predicate collocated with every words in the sentence as argument candidates. Note that since the nominal predicate sometimes takes itself as its argument, the predicate itself is also included in the argument candidate list. 

\begin{table}
	\centering
	\begin{tabular}{lc|lc}
		\hline
		\hline
		Head \quad dependent (word pair) & Label & Head \qquad dependent (word pair) & Label\\
		\hline
		$<$VR$>$ \quad prevent& 01 & $<$VR$>$ \qquad fertilizing & 01\\
		\hline
		prevent \quad The & None & fertilizing \quad The & None\\
		prevent \quad gene & A0 & fertilizing \quad gene & None \\
		prevent \quad thus & AM-MIS & fertilizing \quad thus & None\\
		prevent \quad can & AM-MOD & fertilizing \quad can & None\\
		prevent \quad prevent & None & fertilizing \quad prevent & None\\
		\dots \qquad\quad \dots &\dots & \dots \qquad\qquad \dots &\dots\\
		prevent \quad itself & None & fertilizing \quad itself & A1\\
		\hline
		\hline
	\end{tabular}
	\caption{Word pairs for semantic role label classification.}\label{word_pair}
\end{table}

\section{Model}
\begin{figure}
	\centering
	\includegraphics[scale=0.8]{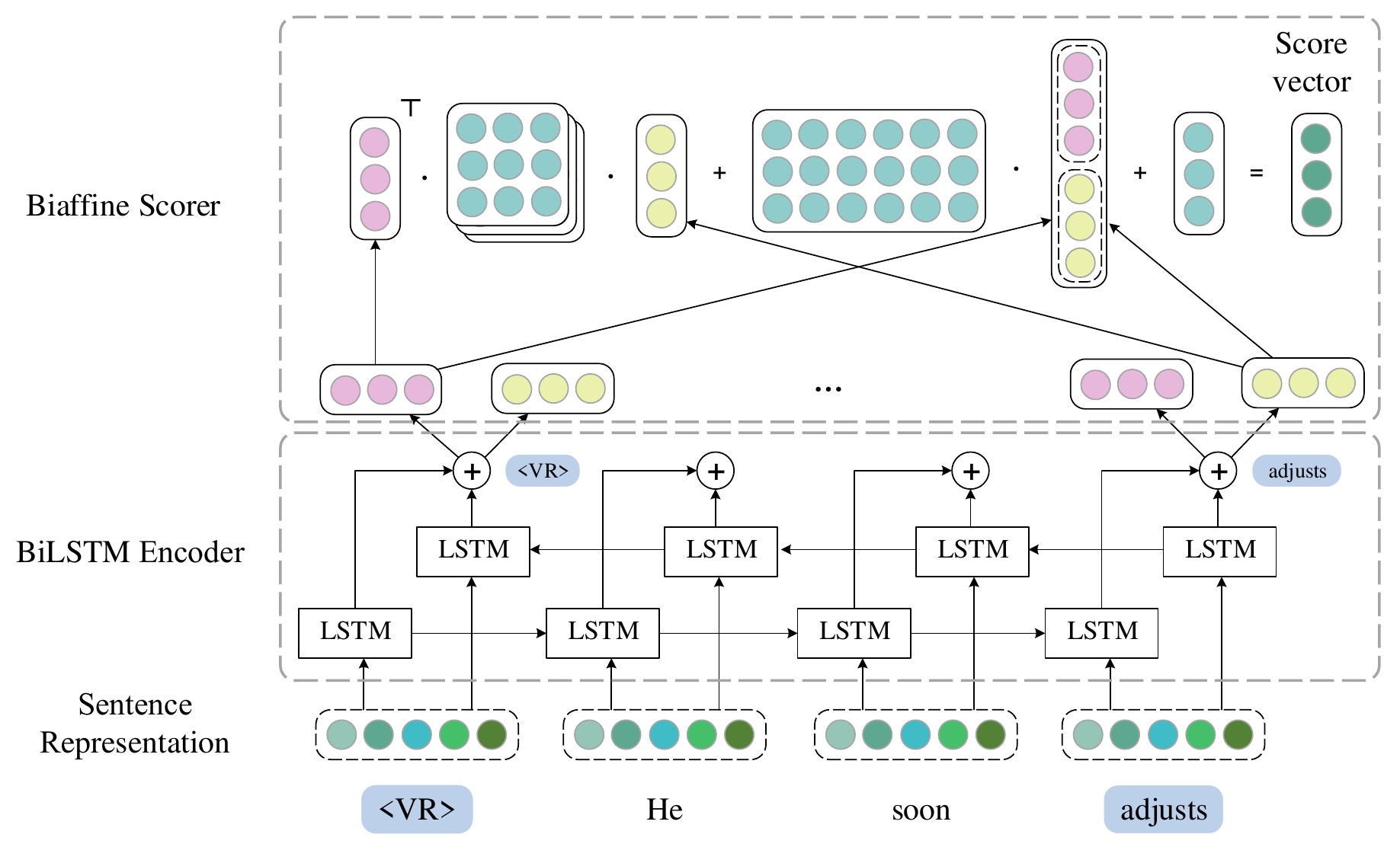}
	\caption{An overview of our model.} \label{biaffine_model}
\end{figure}

Our model contains two main components: (1) a deep BiLSTM encoder that takes each word embedding $\bm{e}$ of the given sentence as input and generates dense vectors for both words in the to-be-classified word pair respectively, (2) a biaffine attentional scorer which takes the hidden vectors for the given word pair as input and predict a label score vector. Figure \ref{biaffine_model} provides an overview of our model.

\subsection{Bidirectional LSTM Encoder}
\paragraph{Word Representation}
The word representation of our model is the concatenation of several vectors: a randomly initialized word embedding $\bm{e}^{(r)}$, a pre-trained word embedding $\bm{e}^{(p)}$, a randomly initialized part-of-speech (POS) tag embedding $\bm{e}^{(pos)}$, a randomly initialized lemma embedding $\bm{e}^{(l)}$. Besides, since previous work \cite{dashimei2018} demonstrated that the predicate-specific feature is helpful in promoting the role labeling process, we employ an indicator embedding $\bm{e}^{(i)}$ to indicate whether a word is a predicate when predicting and labeling the arguments for each given predicate. The final word representation is given by $\bm{e} = \bm{e}^{(r)} \oplus \bm{e}^{(p)} \oplus \bm{e}^{(l)} \oplus \bm{e}^{(pos)} \oplus \bm{e}^{(i)}$, where $\oplus$ is the concatenation operator.

\paragraph{Encoder} 
As commonly used to model the sequential input in most NLP tasks \cite{Wang2016Learning,dashimei2018}, BiLSTM is adopted for our sentence encoder. By incorporating a stack of two distinct LSTMs, BiLSTM processes an input sequence in both forward and backward directions. In this way, the BiLSTM encoder provides the ability to incorporate the contextual information for each word. 

Given a sequence of word representation $S=\{\bm{e}_1, \bm{e}_2, \cdots, \bm{e}_N\}$ as input, the $i$-th hidden state $\bm{g}_i$ is encoded as follows:
\begin{equation*}
\bm{g}^f_i = LSTM^\mathcal{F}\left(\bm{e}_i, \bm{g}^f_{i-1}\right),\ \ \  
\bm{g}^b_i = LSTM^\mathcal{B}\left(\bm{e}_i, \bm{g}^b_{i+1}\right),\ \ \  
\bm{g}_i  = \bm{g}^f_i \oplus \bm{g}^b_i, 
\end{equation*}
where $LSTM^\mathcal{F}$ denotes the forward LSTM transformation and $LSTM^\mathcal{B}$ denotes the backward LSTM transformation. $\bm{g}^f_i$ and $\bm{g}^b_i$ are the hidden state vectors of the forward LSTM and backward LSTM respectively.

\subsection{Biaffine Attentional Role Scorer}

Typically, to predict and label arguments for a given predicate, a role classifier is employed on top of the BiLSTM encoder. Some work like \cite{marcheggiani2017} shows that incorporating the predicate's hidden state in their role classifier enhances the model performance, while we argue that a more natural way to incorporate the syntactic information carried by the predicate is to employ the attentional mechanism. Our model adopts the recently introduced biaffine attention \cite{dozat2017deep} to enhance our role scorer. Biaffine attention is a natural extension of bilinear attention \cite{luong-pham-manning} which is widely used in neural machine translation (NMT). 

\paragraph{Nonlinear Affine Transformation} 
Usually, a BiLSTM decoder takes the concatenation $\bm{g}_i$ of the hidden state vectors as output for each hidden state. However, in the SRL context, the encoder is supposed to distinguish the currently considered predicate from its candidate arguments. To this end, we perform two distinct affine transformations with a nonlinear activation on the hidden state $\bm{g}_i$, mapping it to vectors with smaller dimensionality:
\begin{equation*}
\bm{h}_i^{(pred)} = ReLU \left(\bm{W}^{(pred)}\bm{g}_i + \bm{b}^{(pred)} \right),\ \ 
\bm{h}_i^{(arg)} = ReLU \left(\bm{W}^{(arg)}\bm{g}_i + \bm{b}^{(arg)} \right),
\end{equation*}
where $ReLU$ is the rectilinear activation function, $\bm{h}_i^{(pred)}$ is the hidden representation for the predicate and $\bm{h}_i^{(arg)}$ is the hidden representation for the candidate arguments. 

By performing such transformations over the encoder output to feed the scorer, the latter may benefit from deeper feature extraction. First, ideally, instead of keeping both features learned by the two distinct LSTMs, the scorer is now enabled to learn features composed from both recurrent states together with reduced dimensionality. Second, it provides the ability to map the predicates and the arguments into two distinct vector spaces, which is essential for our tasks since some words can be labeled as predicate and argument simultaneously. Mapping a word into two different vectors can help the model disambiguate the role that it plays in different context. 

\paragraph{Biaffine Scoring} 
In the standard NMT context, given a target recurrent output vector $h_i^{(t)}$ and a source recurrent output vector $h_j^{(s)}$, a bilinear transformation calculates a score $s_{ij}$ for the alignment:
\begin{equation*}
s_{ij} = \bm{h}_i^{\top (t)} \bm{W} \bm{h}_j^{(s)},
\end{equation*}

However, considering that in a traditional classification task, the distribution of classes is often uneven, and that the output layer of the model normally includes a bias term designed to capture the prior probability $P(y_i = c)$ of each class, with the rest of the model focusing on learning the likelihood of each class given the data $P (y_i = c|x_i )$, \cite{dozat2017deep} introduced the bias terms into the bilinear attention to address such uneven problem, resulting in a biaffine transformation. The biaffine transformation is a natural extension of the bilinear transformation and the affine transformation. In SRL task, the distribution of the role labels is similarly uneven and the problem comes worse after we introduce the additional $<$\emph{VR}$>$ node and \emph{None} label, directly applying the primitive form of bilinear attention would fail to capture the prior probability $P(y_i=c_k)$ for each class. Thus, the biaffine attention introduced in our model would be extremely helpful for semantic role prediction.

It is worth noting that in our model, the scorer aims to assign a score for each specific semantic role. Besides learning the prior distribution for each label, we wish to further capture the preferences for the label that a specific predicate-argument pair can take. Thus, our biaffine attention contains two distinguish bias terms: 
\begin{align}
\bm{s}_{ij} =\ & \bm{h}_i^{\top (arg)} \bm{W}^{(role)} \bm{h}_j^{(pred)}\label{bilinear}\\
& + \bm{U}^{(role)}\left( \bm{h}_i^{(arg)}\oplus \bm{h}_j^{(pred)} \right)\label{pair_bias}\\
& + \bm{b}^{(role)} \label{class_bias},
\end{align}
where $\bm{W}^{(role)}$, $\bm{U}^{(role)}$ and $\bm{b}^{(role)}$ are parameters that will be updated by some gradient descent methods in the learning process. There are several points that should be paid attention to in the above biaffine transformation. First, since our goal is to predict the label for each pair of $\bm{h}_i^{(arg)}$, $\bm{h}_j^{(pred)}$, the output of our biaffine transformation should be a vector of dimensionality $N_r$ instead of a real value, where $N_r$ is the number of all the candidate semantic labels. Thus, the bilinear transformation in Eq. (\ref{bilinear}) maps two input vectors into another vector. This can be accomplished by setting $\bm{W}^{(role)}$ as a $(d_h \times N_r \times d_h)$ matrix, where $d_h$ is the dimensionality of the hidden state vector. Similarly, the output of the linear transformation in Eq. (\ref{pair_bias}) is also a vector by setting $\bm{U}^{(role)}$ as a $(N_r \times 2 d_h)$ matrix. Second, Eq. (\ref{pair_bias}) captures the preference of each role (or sense) label condition on taking the $j$-th word as predicate and the $i$-th word as argument. Third, the last term $\bm{b}^{(role)}$ captures the prior probability of each class $P(y_i = c_k)$. Notice that Eq. (\ref{pair_bias}) and (\ref{class_bias}) capture different kinds of bias for the latent distribution of the label set.

Given a sentence of length $L$ (including the $<$\emph{VR}$>$ node), for one of its predicates $w_j$, the scorer outputs a score vector $\{\bm{s}_{1j}, \bm{s}_{2j}, \cdots, \bm{s}_{Lj}\}$. Then our model picks as its output the label with the highest score from each score vector: $y_{ij}=\mathop{\arg\max}_{1\leq k\leq N_r} (\bm{s}_{ij}[k])$, where $\bm{s}_{ij}[k]$ denotes the score of the $k$-th candidate semantic label.

\section{Experiments}
\subsection{Dataset and Training Detail}\label{hyperparameters}
We evaluate our model\footnote{The code is available at \url{https://github.com/JiaxunCai/Dynet-Biaffine-SRL}} on English and Chinese CoNLL-2009 datasets with the standard split into training, test and development sets. The pre-trained embedding for English is trained on Wikipedia and Gigaword using the GloVe \cite{pennington2014glove}, while those for Chinese is trained on Wikipedia. Our implementation uses the DyNet\footnote{https://github.com/clab/dynet} library for building the dynamic computation graph of the network. 

When not otherwise specified, our model uses: $100$-dimensional word, lemma, pre-trained and POS tag embeddings and $16$-dimensional predicate-specific indicator embedding; and a $20\%$ chance of dropping on the whole word representation; $3$-layer BiLSTMs with $400$-dimensional forward and backward LSTMs, using the form of recurrent dropout suggested by \cite{Gal2016Dropout} with an $80\%$ keep probability between time-steps and layers; two $300$-dimensional affine transformation with the ReLU non-linear activation on the output of BiLSTM, also with an $80\%$ keep probability. 

The parameters in our model are optimized with Adam \cite{Kingma2015Adam}, which keeps a moving average of the L2 norm of the gradient for each parameter throughout training and divides the gradient for each parameter by this moving average, ensuring that the magnitude of the gradients will on average be close to one. For the parameters of optimizer, we follow the settings in \cite{dozat2017deep}, with $\beta_1 = \beta_2 = 0.9$ and learning rate $0.002$, annealed continuously at a rate of $0.75$ every $5,000$ iterations, with batches of approximately $5,000$ tokens. The maximum number of epochs of training is set to $50$.

\subsection{Results}
Tables \ref{eng_tbl} and \ref{chn_rest} report the comparison of performance between our model and previous dependency-based SRL model on both English and Chinese. Note that the predicate disambiguation subtask is unifiedly tackled with arguments labeling in our model with precisions of 95.0\% and 95.6\% respectively on English and Chinese test sets in our experiments\footnote{Note that we give comparable predicate disambiguation results with \newcite{dashimei2018}, with 95.01\% and 95.58\% F$_1$ on development and test sets, respectively.}. The proposed model accordingly outperforms all the SRL systems so far on both languages, even including those syntax-aware and ensemble ones. The improvement grows even larger when comparing only with the single syntax-agnostic models. 

For English, our syntax-agnostic model even slightly outperforms the best reported syntax-aware model \cite{dashimei2018} with a margin of 0.1\% F$_1$. Compared to syntax-agnostic models, our model overwhelmingly outperforms (with an improvement of 0.9\% F$_1$) the previous work \cite{dashimei2018}.

\begin{table}
\centering
\begin{tabular}{lccc}
	\hline
	
	\hline
	\textit{Syntax-aware system (single)} & P & R & F$_1$ \\
	\hline
	\newcite{Zhao2009Conll} & $-$ & $-$ & 86.2\\
	\newcite{Zhao2009Conll-SRL} & $-$ & $-$ & 85.4 \\
	\newcite{bjorkelund2010} & 87.1 & 84.5 & 85.8 \\
	\newcite{Lei2015} & $-$ & $-$ & 86.6 \\
	\newcite{Fitzgerald2015} & $-$ & $-$ & 86.7 \\
	\newcite{roth2016} & 88.1 & 85.3 & 86.7\\
	\newcite{marcheggianiEMNLP2017} & 89.1 & 86.8 & 88.0 \\
	\textbf{\newcite{dashimei2018}} & \textbf{89.7} & \textbf{89.3} & \textbf{89.5} \\
	\hline
	\textit{Syntax-aware system (ensemble)} & P & R & F$_1$ \\
	\hline
	\newcite{roth2016} & 90.3 & 85.7 & 87.9\\
	\newcite{marcheggianiEMNLP2017} & 90.5 & 87.7 & 89.1 \\
	\hline
	\hline
	\textit{Syntax-agnostic system} & P & R & F$_1$ \\
	\hline
	\newcite{marcheggiani2017} & 88.7 & 86.8 & 87.7 \\
	\newcite{dashimei2018} & 89.5 & 87.9 & 88.7 \\
	\textbf{This work} & \textbf{89.9} & \textbf{89.2} & \textbf{89.6} \\
	\hline
	
	\hline
\end{tabular}
\caption{Results on English in-domain (WSJ) test set.}\label{eng_tbl}
\end{table}

\begin{table}
\setlength{\tabcolsep}{3.5pt}
\begin{minipage}[t]{0.48\linewidth}
	\vspace{0pt}
	\centering
	\begin{tabular}{lccc}
		\hline
		
		\hline
		\textit{Syntax-aware system} & P & R & F$_1$ \\
		\hline
		\newcite{Zhao2009Conll} & 80.4 & 75.2 & 77.7 \\
		\newcite{bjorkelund2009} & 82.4 & 75.1 & 78.6 \\
		\newcite{roth2016} & 83.2 & 75.9 & 79.4 \\
		\newcite{marcheggianiEMNLP2017} & 84.6 & 80.4 & 82.5 \\
		\textbf{\newcite{dashimei2018}} & \textbf{84.2} & \textbf{81.5} & \textbf{82.8} \\
		\hline
		\hline
		\textit{Syntax-agnostic system} & P & R & F$_1$ \\
		\hline
		\newcite{marcheggiani2017} & 83.4 & 79.1 & 81.2 \\
		\newcite{dashimei2018} & 84.5 & 79.3 & 81.8 \\
		\textbf{This work} & \textbf{84.7} & \textbf{84.0} & \textbf{84.3} \\
		\hline
		
		\hline
	\end{tabular}
	\caption{Results on Chinese in-domain test set.}\label{chn_rest}
\end{minipage}
\hspace{0.02\linewidth}
\begin{minipage}[t]{0.48\linewidth}
	\vspace{0pt}
	\centering
	\begin{tabular}{lccc}
		\hline
		
		\hline
		\textit{Syntax-aware system} & P & R & F$_1$ \\
		\hline
		\newcite{bjorkelund2010} & 75.7 & 72.2 & 73.9 \\
		\newcite{Lei2015} & $-$ & $-$ & 75.6 \\
		\newcite{Fitzgerald2015} & $-$ & $-$ & 75.2 \\
		\newcite{roth2016} & 76.9 & 73.8 & 75.3 \\
		\newcite{marcheggianiEMNLP2017} & 78.5 & 75.9 & 77.2 \\
		\textbf{\newcite{dashimei2018}} & \textbf{81.9} & \textbf{76.9} & \textbf{79.3} \\
		\hline
		\hline
		\textit{Syntax-agnostic system} & P & R & F$_1$ \\
		\hline
		\newcite{marcheggiani2017} & 79.4 & 76.2 & 77.7 \\
		\newcite{dashimei2018} & 81.7 & 76.1 & 78.8 \\
		\textbf{This work} & \textbf{79.8} & \textbf{78.3} & \textbf{79.0} \\
		\hline
		
		\hline
	\end{tabular}
	\caption{Results on English out-of-the-domain (Brown) test set.}\label{ood_tbl}
\end{minipage}
\end{table}

Although we used the same parameters as for English, our model substantially outperforms the state-of-art models on Chinese, demonstrating that our model is robust and less sensitive to the parameter selection. For Chinese, the proposed model outperforms the best previous model \cite{dashimei2018} with a considerable improvement of 1.5\% F$_1$, and surpasses the best single syntax-agnostic model \cite{dashimei2018} with a margin of 2.5\% F$_1$. 

Table \ref{ood_tbl} compares the results on English out-of-the-domain (Brown) test set, from which our model still remains strong. The proposed model gives a comparable result with the highest score from syntax-aware model of \cite{dashimei2018}, which affirms that our model does well learn and generalize the latent semantic preference of the data. 

\begin{table}
	\centering
	\begin{tabular}{l|ccc|c}
		\hline
		
		\hline
		& \multicolumn{3}{|c|}{\textbf{AL}} & \textbf{PD}\\
		System & P & R & F$_1$ & P\\
		\hline
		without POS & 89.5 & 89.1 & 89.3 & 94.9\\
		without lemma & 89.5 & 89.3 & 89.4 & 94.9\\
		without indicator & 89.1 & 88.5 & 88.8 & 95.0\\
		\textbf{This work} & \textbf{89.9} & \textbf{89.2} & \textbf{89.6} & \textbf{95.0}\\
		\hline
		
		\hline
	\end{tabular}
	\caption{Contribution of the input representation. Acronyms used: \textbf{AL}-argument labeling, \textbf{PD}-predicate disambiguation.}\label{input_abl_tbl}
\end{table}

Results on both in-domain and out-of-the-domain test sets demonstrate the effectiveness and the robustness of the proposed model structure---the non-linear transformation after the BiLSTM serves to distinguish the predicate from argument while the biaffine attention tells what to attend for each candidate argument. In Section \ref{insight}, we will get an insight into our model and explore how each individual component impacts the model performance.

\subsection{Ablation Analysis}
\subsubsection{Word Representation}
To learn how the input word representation choice impacts our model performance, we conduct an ablation study on the English test set whose results are shown in Table \ref{input_abl_tbl}. Since we deal with the two subtasks in a single model, the choice of word representation will simultaneously influence the results of both of them. Besides the results of argument labeling, we also report the precision of predicate disambiguation. 
 
The results demonstrate that the multiple dimensional indicator embedding proposed by (He et al., 2018) contributes the most to the final performance of our model. It is consistent with the conclusion in \cite{marcheggiani2017} which argue that encoding predicate information promotes the SRL model. It is interesting that the impact of POS tag embedding (about 0.3\% F$_1$) is less compared to the previous works, which possibly allows us to build an accuracy model even when the POS tag label is unavailable. 

\subsubsection{Into the Model}\label{insight}
In this section, we get insight into the proposed model, exploring how the deep BiLSTM encoder and the biaffine attention affect the labeling results respectively. Specifically, we present two groups of results on the CoNLL-2009 English test set. 1) Shallow biaffine attentive (SBA) labeler. Instead of mapping the output of the BiLSTM into two distinct vector spaces, we apply a single non-linear affine transformation on the output. The single transformation just serves to reduce the dimensionality and does not differ the predicates from the arguments. 2) Deep bilinear attentive (DBA) labeler. We apply the primitive form of bilinear attention in the scorer by removing the two bias terms of the biaffine transformation. By this means, we learn to what extent can the bias terms fit the prior distribution of the data. Results of the two experiments are shown in Table \ref{component_abl_tbl}. 

\begin{table}
\begin{minipage}[t]{0.48\linewidth}
	\vspace{0pt}
	\centering
	\begin{tabular}{l|ccc|c}
		\hline
		
		\hline
		& \multicolumn{3}{|c|}{\textbf{AL}} & \textbf{PD}\\
		System & P & R & F$_1$ & P\\
		\hline
		SBA-labeler & 89.5 & 88.8 & 89.1 & 94.7\\
		DBA-labeler & 88.1 & 87.7 & 87.9 & 94.3\\
		\textbf{This work} & \textbf{89.9} & \textbf{89.2} & \textbf{89.6} & \textbf{95.0}\\
		\hline
		
		\hline
	\end{tabular}
	\caption{Contribution of the model components.}\label{component_abl_tbl}
\end{minipage}
\hspace{0.02\linewidth}
\begin{minipage}[t]{0.48\linewidth}
	\vspace{0pt}
	\centering
	\begin{tabular}{l|ccc|c}
		\hline
		
		\hline
		& \multicolumn{3}{|c|}{\textbf{AL}} & \textbf{PD}\\
		System & P & R & F$_1$ & P\\
		\hline
		\emph{with} pruning & 88.2 & 85.6 & 86.9 & 95.0\\
		\emph{without} pruning & \textbf{89.9} & \textbf{89.2} & \textbf{89.6} & \textbf{95.0} \\
		\hline
		
		\hline
	\end{tabular}
	\caption{Comparison of results with and without argument candidate pruning.}\label{pruning_tbl}
\end{minipage}
\end{table}

The results show that the bias terms in biaffine attention play an important role in promoting the model performance. Removal of the bias terms dramatically declines the performance by 1.7\% F$_1$. Thus we can draw a conclusion that the bias term does well in fitting the prior distribution and global preference of the data. The bilinear attentional model behaves more poorly since it struggles to learn the likelihood of each class on an uneven data set without knowledge about the prior distribution.
Though the deep encoder contributes less to the performance, it also brings an improvement of 0.5\% F$_1$. Note that the only difference of SBA-labeler of our standard model is whether the hidden representations of the arguments and the predicates lay in different vector spaces. Such a result confirms that distinguishing the predicates from the arguments in encoding process indeed enhances the model to some extent.

\subsubsection{Syntax-aware and Syntax-agnostic} Noting that the work \cite{Zhao2008Parsing} and \cite{zhao-jair-2013} are similar to ours in modeling the dependency-based SRL tasks as word pair classification, and that they successfully incorporate the syntactic information by applying argument candidate pruning, we further perform empirical study to explore whether employing such pruning method enhance or hinder our model. Specifically, we use the automatically predicted parse with moderate performance provided by CoNLL-2009 shared task, with the LAS score about 86\%.

\begin{figure}
\begin{minipage}[t]{0.475\linewidth}
	\centering
	\includegraphics[scale=0.7]{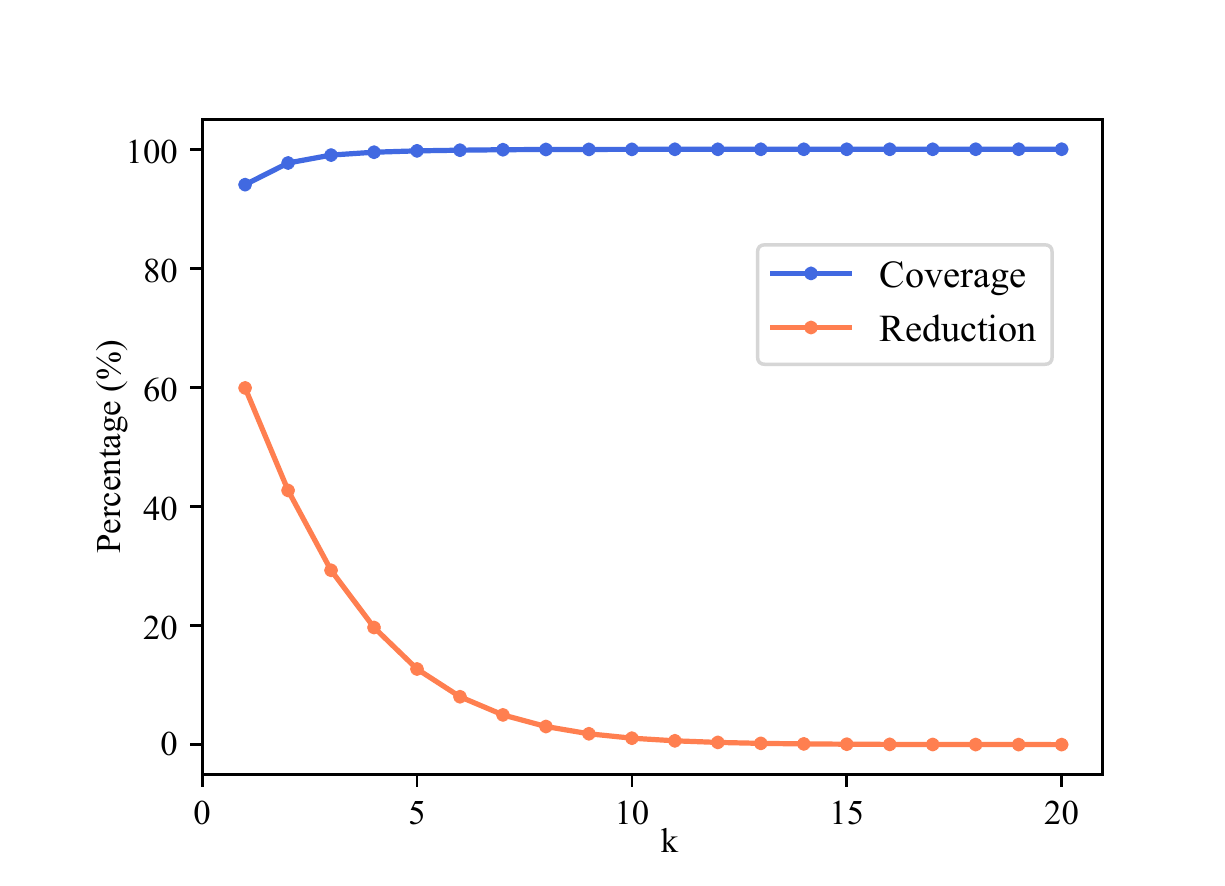}
	\caption{Coverage rate of true arguments and reduction rate of argument candidates against the pruning order $k$ on the English training set.}\label{k_coverage}
\end{minipage}
\hspace{0.04\linewidth}
\begin{minipage}[t]{0.475\linewidth}
	\centering
	\includegraphics[scale=0.5]{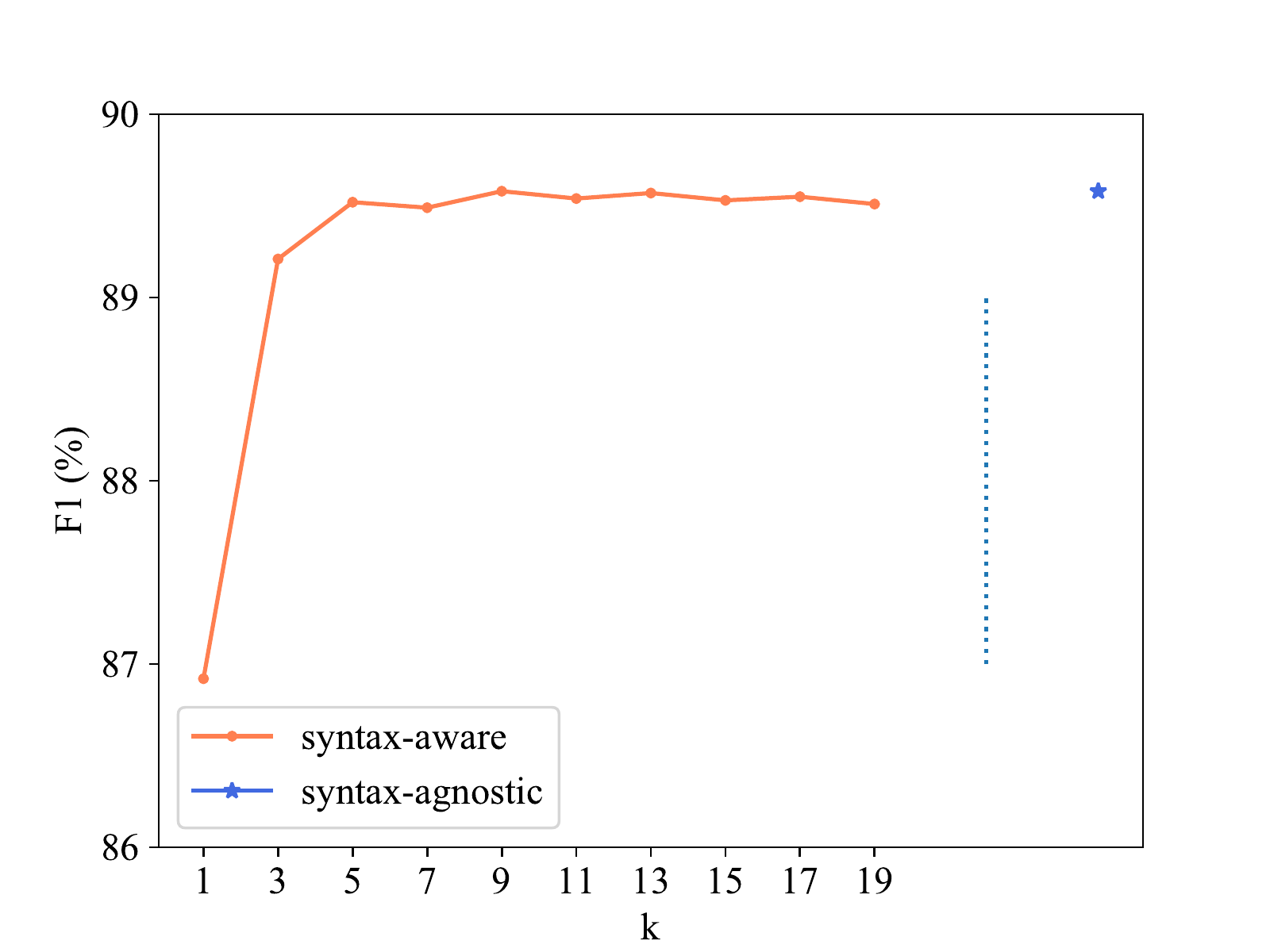}
	\caption{F1 scores against the pruning order $k$ on English test set.}\label{k_pruning}
\end{minipage}
\end{figure}

 The pruning method is supposed to work since it can alleviate the imbalanced label distribution caused by introducing the \emph{None} label. However, as shown in Table \ref{pruning_tbl}, the result is far from satisfying. The main reason might be the pruning algorithm is so strict that too many true arguments are falsely pruned. To address this problem, \newcite{dashimei2018} introduced an extended $k$-order argument pruning algorithm. Figure \ref{k_coverage} shows the curves of coverage and reduction rate against the pruning order $k$ on the English training set following \cite{dashimei2018}. Following this work, we further perform different orders of pruning and obtain the F$_1$ scores curve shown in Figure \ref{k_pruning}. However, the $k$-order pruning does not boost the performance of our model. Table \ref{syntax_gap} presents the performance gap between syntax-agnostic and syntax-aware settings of the same models. Unlike the other two works, the introduction of syntax information fails to bring about bonus for our model. Nevertheless, it is worth noting that even when running without the syntax information, our mode still show a promising result compared to the other syntax-aware models.
 
\begin{table}
	\centering
	\begin{tabular}{lccc}
		\hline
		
		\hline
		System & syntax-agnostic & syntax-aware & $\Delta$ F$_1$ \\
		\hline
		\newcite{marcheggianiEMNLP2017} & 87.7 & 88.0 & 0.3\\
		
		\newcite{dashimei2018} (CoNLL-2009 predicted) & 88.7 & 89.5 & 0.8\\
		\newcite{dashimei2018} (gold syntax) & 88.7 & 90.3 & 1.6\\
		\hline
		This work (CoNLL-2009 predicted) & 89.6 & 89.6 & $\approx$0\\
		\hline
		
		\hline
	\end{tabular}
	\caption{The absolute performance gaps between syntax-agnostic and syntax-aware settings. Both \cite{dashimei2018} and our models use 10-order pruning according to syntactic parse tree.}\label{syntax_gap}
\end{table}

\subsection{CoNLL 2008: Augment the Model with Predicate Identification}\label{ablsec_conll08}
Though CoNLL-2009 provided the gold predicate beforehand, the predicate identification subtask is still indispensable for a real world SRL task. Thus, we further augment our model with the predicate identification ability. 

Specifically, we first attach all the words in the sentence to the virtual root $<$\emph{VR}$>$ and label the word which is not a predicate with the \emph{None} role label. It should be noting that, in CoNLL-2009 settings, we just attach the predicates to the virtual root, since we do not need to distinguish the predicate from other word. The training scheme still keeps the same as that in CoNLL-2009 settings, while in testing phase, an additional procedure is performed to find out all the predicates of a given sentence.

First, our model is fed the representations of the virtual root and each word of the input sentence, identifying and disambiguating all the predicates of the sentence. Second, it picks each predicate predicted by the model with each word of the sentence to identify and label the semantic role in between, which remains the same as the model does on CoNLL-2009. The second phase is repeated until all the predicates have got its arguments being identified and labeled. We evaluate our model on CoNLL-2008 benchmark using the same hyperparameters settings mentioned in Section \ref {hyperparameters} except that we remove the predicate-specific indicator feature. 

The F$_1$ scores on predicates identification and labeling of our model is 89.43\%, which remain comparable with the most recent work \cite{dashimei2018} (90.53\% F$_1$). As shown in Table \ref{conll08_results}, though tackling all the subtasks of CoNLL-2008 SRL unifiedly in a full end-to-end manner, our model outperforms the best reported results with a large margin of about 1.7\% semantic F$_1$. 
 
\begin{table}[!htb]
\setlength{\tabcolsep}{4.6pt}
\centering
\begin{tabular}{l|c|ccc|ccc}
	\hline
		
	\hline
	& & \multicolumn{3}{c|}{\textbf{Predicate Labeling}} & \multicolumn{3}{c}{\textbf{Semantic Labeling}} \\
	System & LAS & P & R & F$_1$ & P & R & F$_1$ \\
	\hline
	\newcite{Johansson2008Dependency} & 90.13 & $-$ & $-$ & $-$ & $-$ & $-$ & 81.75 (80.37) \\
	\newcite{Zhao2008Parsing} & 87.52 & $-$ & $-$ & $-$ & 80.57 & 74.97 & 77.67 \\
	\newcite{zhao2009} & 88.39 & $-$ & $-$ & $-$ & $-$ & $-$ & 82.1 (80.53) \\
			& 89.28 & $-$ & $-$ & $-$ & $-$ & $-$ & 82.5 (80.94) \\
	\newcite{zhao-jair-2013} & 88.39 & $-$ & $-$ & (87.15) & $-$ & $-$ & 82.5 (80.91) \\
			& 89.28 & $-$ & $-$ & (86.47) & $-$ & $-$ & 82.4 (80.88) \\
	\newcite{dashimei2018} (syntax-aware) & 86.0 & 89.73 & 91.35 & 90.53  & 83.9 & 82.7 & 83.3 \\
	\hline
	\newcite{dashimei2018} (syntax-agnostic) & $-$ & 89.73 & 91.35 & 90.53 & 83.5 & 82.4 & 82.9 \\
	\textbf{Ours (syntax-agnostic)} & $-$ & \textbf{88.9} & \textbf{90.0} & \textbf{89.4 (87.9)} & \textbf{84.7} & \textbf{85.2} & \textbf{85.0 (83.6)} \\
	\hline
			
	\hline
\end{tabular}
\caption{Results on the CoNLL-2008 test set (WSJ). The results enclosed with parenthesis are evaluated on WSJ + Brown test set, following the official evaluation setting of CoNLL-2008 shared task.}\label{conll08_results}
\end{table}

\section{Related Work}
Semantic role labeling was pioneered by \newcite{gildea2002}. Most traditional SRL models heavily rely on complex feature engineering \cite{pradhan2005,Zhao2009Conll,bjorkelund2009}. Among those early works, \newcite{pradhan2005} combined features derived from different syntactic parses based on SVM classifier, while \newcite{Zhao2009Conll} exploited the abundant set of language-specific features that were carefully designed for SRL task. 

In recent years, applying neural networks in SRL task has gained a lot of attention due to the impressive success of deep neural networks in various NLP tasks \cite{Zhang2016Probabilistic,Cai2017Fast,Qin2017Adversarial,Cai2017Pair}. \newcite{Collobert2011} initially introduced neural networks into the SRL task. They developed a feed-forward network that employed a convolutional network as sentence encoder and a conditional random field as a role classifier. \newcite{Foland2015} extended their model to further use syntactic information by including binary indicator features. \newcite{Fitzgerald2015} exploited a neural network to unifiedly embed arguments and semantic roles, similar to the work \cite{Lei2015} which induced a compact feature representation applying tensor-based approach. \newcite{roth2016} introduced the dependency path embedding to incorporate syntax and exhibited a notable success, while \newcite{marcheggianiEMNLP2017} employed the graph convolutional network to integrate syntactic information into their neural model.

Besides the above-mentioned works who relied on syntactic information, several works attempted to build SRL systems without or with little syntactic information. \newcite{zhou-xu2015} came up with an end-to-end model for span-based SRL and obtained surprising performance putting syntax aside. \newcite{he-acl2017} further extended their work with the highway network. Simultaneously, \newcite{marcheggiani2017} proposed a syntax-agnostic model with effective word representation for dependency-based SRL.

However, almost all of previous works treated the predicate disambiguation as individual subtasks, apart from \cite{Zhao2008Parsing,Zhao2009Conll,Zhao2009Conll-SRL,zhao-jair-2013}, who presented the first end-to-end system for dependency SRL. For the neural models of dependency SRL, we have presented the first end-to-end solution that handles both semantic labeling subtasks in one single model. At the same time, our model enjoys the advantage that does not rely any syntactic information.

This work is also closely related to the attentional mechanism. The traditional attention mechanism was proposed by \newcite{Bahdanau15} in the NMT literature. Following the work \cite{luong-pham-manning} that encouraged substituting the MLP in the attentional mechanism with a single bilinear transformation, \newcite{dozat2017deep} introduced the bias terms into the primitive form of bilinear attention and applied it for dependency parsing. They demonstrate that the bias terms help their model to capture the uneven prior distribution of the data, which is again verified by our practice on SRL in this paper.

Different from the latest strong syntax-agnostic models in \cite{marcheggianiEMNLP2017} and \cite{dashimei2018} which both adopted sequence labeling formulization for the SRL task, this work adopts word pair classification scheme implemented by LSTM encoder and biaffine scorer. Compared to the previous state-of-the-art syntax-agnostic model in \cite{dashimei2018} whose performance boosting (more than 1\% absolute gain) is mostly due to introducing the enhanced representation, namely, the CNN-BiLSTM character embedding from \cite{ELMo}, our performance promotion mainly roots from model architecture improvement, which results in quite different syntax-aware enhanced impacts. Using the same latest syntax-aware $k$-order pruning, the syntax-agnostic backbone in \cite{dashimei2018} may receive about 1\% performance gain, while our model is furthermore enhanced little. This comparison also suggests the possibility that maybe our model can be further improved by incorporating with the same character embedding as \cite{dashimei2018} does\footnote{Such an attempt may be hindered by too luxurious computational resource requirement, as there comes extremely high graphic memory prerequisite when integrating both biaffine scorer and the ELMo character embedding.}.

\section{Conclusion and Future Work}
This paper presents a full end-to-end neural model for dependency-based SRL. It is the first time that a SRL model shows its ability to unifiedly handle the predicate disambiguation and the argument labeling subtasks. Our model is effective while remains simple. Experiments show that it achieves the best scores on CoNLL benchmark both for English and Chinese, outperforming the previous state-of-the-art models even with syntax-aware features. Our further investigation by incorporating the latest syntax-aware pruning algorithm shows that the proposed model is insensitive to the input syntactic information, demonstrating an interesting performance style for the SRL task. Of course, we cannot exclude the possibility that the proposed model can be furthermore improved by other syntactic information integration ways, which is left for the future work.

\bibliographystyle{acl}
\bibliography{acl2018}

\end{document}